\documentclass[autodetect-engine,11pt,a4paper]{article}
\usepackage{tikz}

\usepackage[hyperref]{emnlp2018}
\usepackage{times}
\usepackage{latexsym}
\usepackage{amsmath}
\usepackage{url}
\usepackage{graphicx}
\usepackage{booktabs}
\usepackage{subfigure}
\usepackage{caption}

\aclfinalcopy

\title{\textbf{Machine Translation Evaluation with BERT Regressor}}

\author{Hiroki Shimanaka$^{\dagger}$ \\\And
  Tomoyuki Kajiwara$^{\dagger\ddagger}$ \\
  $^{\dagger}$Graduate School of Systems Design, Tokyo Metropolitan University, Tokyo, Japan \\
  {\tt shimanaka-hiroki@ed.tmu.ac.jp, komachi@tmu.ac.jp} \\
  $^{\ddagger}$Institute for Datability Science, Osaka University, Osaka, Japan \\
  {\tt kajiwara@ids.osaka-u.ac.jp} \\\And
  Mamoru Komachi$^{\dagger}$ \\
}

\date{}
\begin{document}
\maketitle

\begin{abstract}
We introduce the metric that using BERT (Bidirectional Encoder Representations from Transformers)~\cite{devlin-2019} for automatic machine translation evaluation.
% BERTは大規模な生コーパスで事前学習した後に少量のデータセットでfine-tuningすることにより、様々なタスクで最高性能を収めている。
% 我々は事前学習されたBERTを人手評価値付きのデータセットでfine-tuningすることにより翻訳文の評価を行う。
The experimental results of the WMT-2017 Metrics Shared Task dataset show that our metric achieves a state-of-the-art performance in segment-level metrics task for all to-English language pairs.
\end{abstract}

\section{Introduction}\label{sec:intro}
% 本研究では，参照文を用いた文単位での機械翻訳自動評価手法について述べる．
% 人手評価との相関が高い文単位の評価ができることにより機械翻訳システムの細かい改善が可能になる．
This study describes a segment-level metric for automatic machine translation evaluation (MTE).
The MTE metrics with a high correlation with human evaluation enable the continuous integration and deployment of a machine translation (MT) system.

% 我々の先行研究~\cite{shimanaka-2018b}では，単語N-gramなどの局所的な素性に基づく従来手法~\cite{ma-2017}では扱えない大域的な情報を考慮するために，大規模コーパスによって事前学習された文の分散表現を用いる機械翻訳自動評価手法RUSE\footnote{\url{https://github.com/Shi-ma/RUSE}} (Regressor Using Sentence Embeddings)を提案した．
% RUSEは，機械翻訳自動評価手法の性能を競うWMT-2018 Metrics Shared Task~\cite{ma-2018}において，文単位の全てのto-English言語対で最高性能を達成した．
% この結果は，事前学習された文の分散表現が機械翻訳の自動評価にとって有用な素性であることを示す．
In our previous study~\cite{shimanaka-2018b}, we proposed RUSE\footnote{\url{https://github.com/Shi-ma/RUSE}}  (Regressor Using Sentence Embeddings) that is a segment-level MTE metric using pre-trained sentence embeddings capable of capturing global information that cannot be captured by local features based on character or word N-grams.
In WMT-2018 Metrics Shared Task~\cite{ma-2018}, RUSE was the best metric on segment-level for all to-English language pairs.
This result indicates that pre-trained sentence embeddings are effective feature for automatic evaluation of machine translation.

% このような文単位での表現学習に関する研究は近年急速に発展している．
% 特に，BERT (Bidirectional Encoder Representations from Transformers)~\cite{devlin-2019}が多くの応用タスクで最高性能を更新し，注目を集めている．
% BERTは，大規模な生コーパスを用いて双方向言語モデルおよび隣接文推定の事前学習を行った上で，タスクに応じた再訓練を行う．
% 例えば，極性分類のような単一文の分類タスクを解く場合と含意関係認識のような文対の分類タスクを解く場合では，異なる方法で再訓練を行う．
% これによって，機械翻訳自動評価の類似タスクである文対の意味的類似度推定タスクにおいても，高い性能を発揮している．
Research related to applying pre-trained language representations to downstream tasks has been rapidly developing in recent years.
In particular, BERT (Bidirectional Encoder Representations from Transformers)~\cite{devlin-2019} has achieved the best performance in many downstream tasks and is attracting attention.
BERT is designed to pre-train using ``masked language model" (MLM) and ``next sentence prediction" (NSP) on large amounts of raw text and fine-tune for a supervised downstream task.
For example, in the case of solving single sentence classification tasks such as sentiment analysis and in the case of solving sentence-pair classification tasks such as  natural language inference task, fine-tuning is performed in different ways.
As a result, BERT also performs well in the task of estimating the similarity between sentence pairs, which is considered to be a similar task of automatic machine translation evaluation.

% そこで本研究では，BERTを用いた機械翻訳の自動評価を行う．
% WMT-2017 Metrics Shared Task~\cite{bojar-2017}のデータセットにおける実験の結果，BERTは文単位の全てのto-English言語対でRUSEを凌ぎ，最高性能を更新した．
% 詳細な分析の結果，RUSEとの主な相違点である{\bf 事前学習の方法}，{\bf 文対モデリング}，{\bf 符号化器の再訓練}の3点が，それぞれBERTの性能改善に貢献していることが明らかになった．
Therefore, we propose the MTE metric that using BERT.
The experimental results in segment-level metrics task conducted using the datasets for all to-English language pairs on WMT17 indicated that the proposed metric shows higher correlation with human evaluations than RUSE, and achieves the best performance.
As a result of detailed analysis, it is clarified that the three main points of difference with RUSE, \textit{the pre-training method}, \textit{the sentence-pair encoding}, and \textit{the fine-tuning of the pre-trained encoder}, contribute to the performance improvement of BERT.

\begin{figure*}[t]
  \centering
  \subfigure[MTE with RUSE.]{\includegraphics[width=0.32\paperwidth]{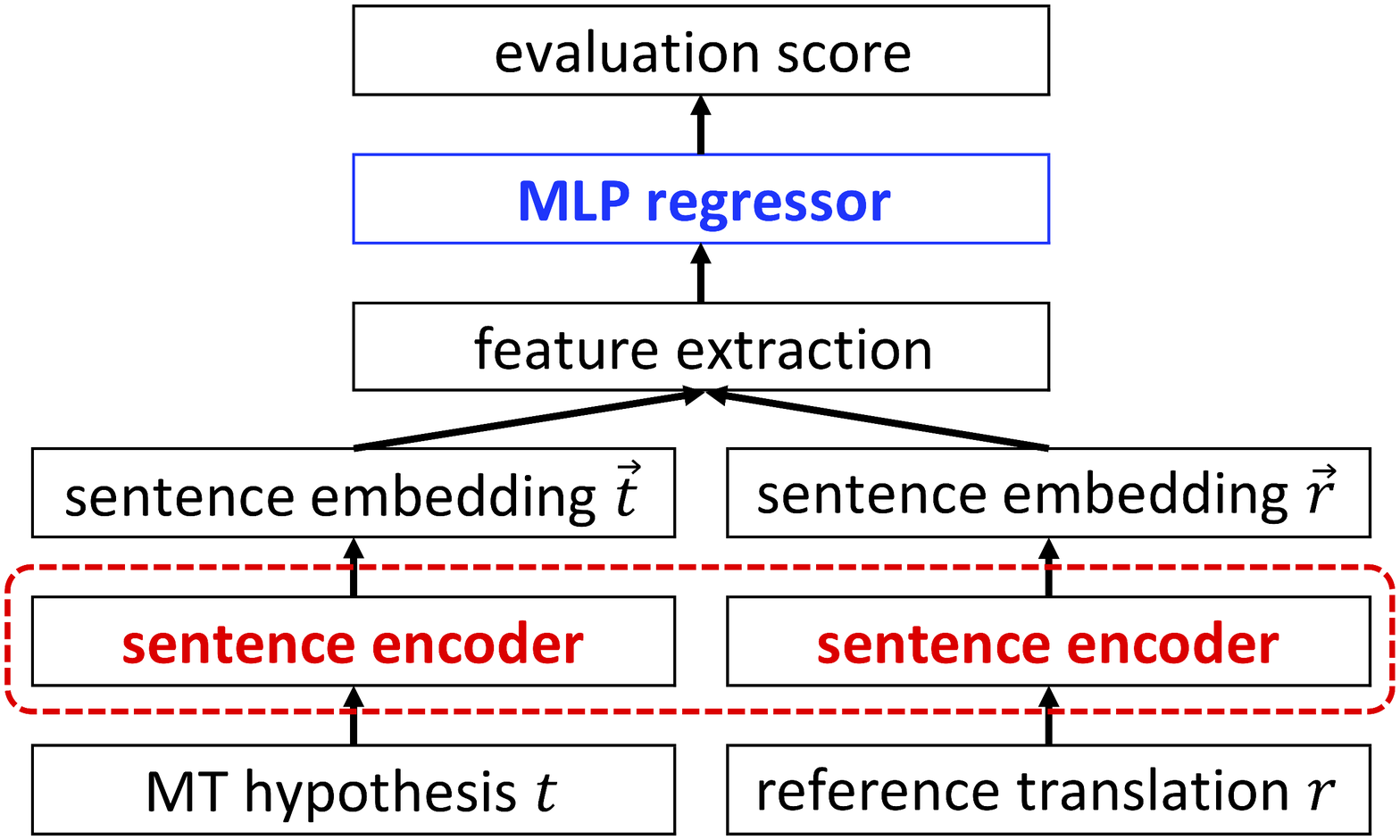}\label{fig:ruse}}
  \hspace{5mm}
  \subfigure[MTE with BERT.]{\includegraphics[width=0.32\paperwidth]{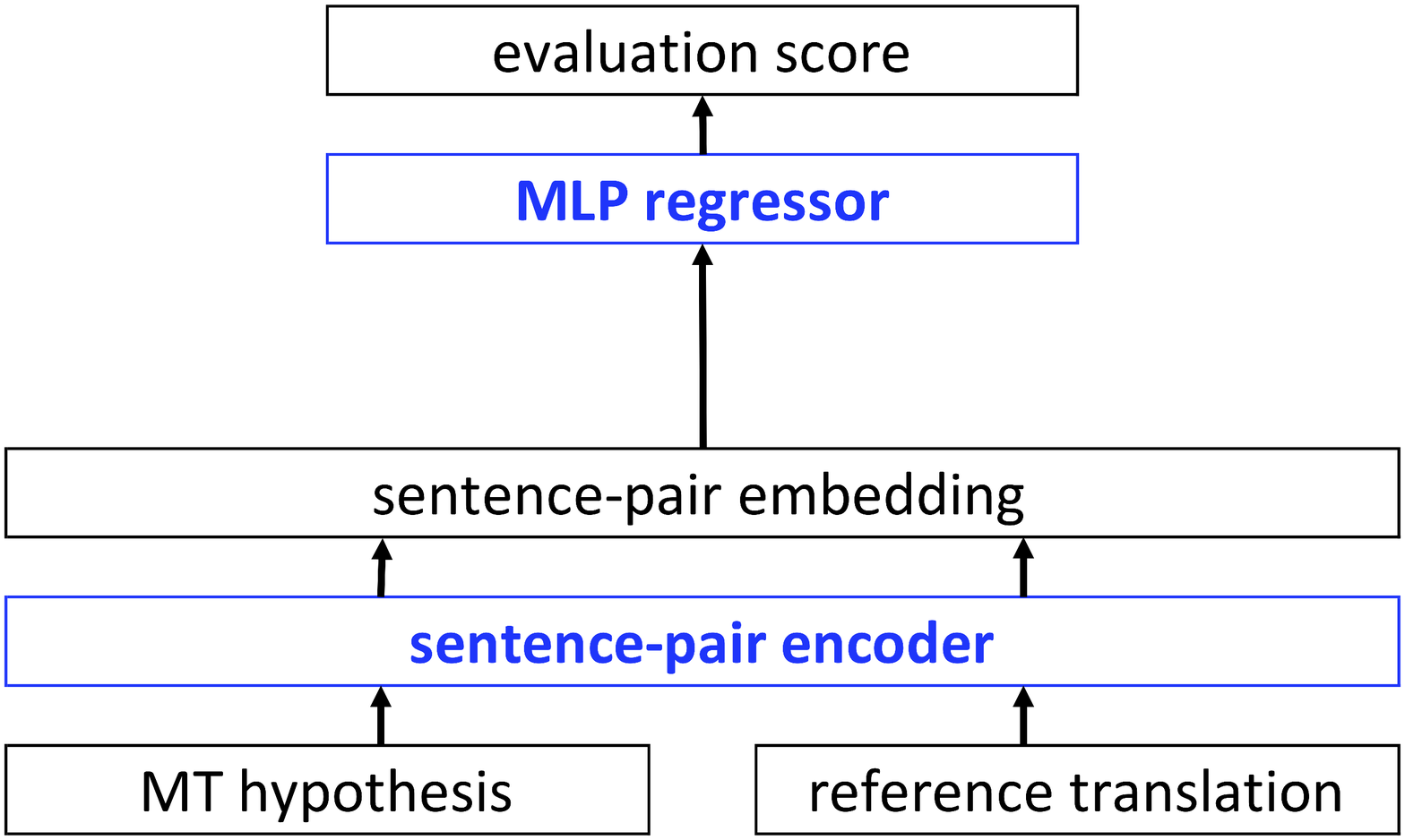}\label{fig:bert}}
 \caption{Outline of each metric. Blue is training but red is fixed.}
\end{figure*}

\section{Related Work}\label{sec:previous_work}
% 本節では，WMT-2017~\cite{bojar-2017}および-2018~\cite{ma-2018}のMetrics Shared Taskにおいて最高性能を達成した機械翻訳自動評価手法について説明する．
% 本タスクでは，機械翻訳の翻訳文に対して人手で参照文および評価値が付与されたデータセットを利用する．
% 各手法は，翻訳文と参照文の文対を入力として評価値を推定し，人手評価とのピアソンの相関係数によって評価される．
% 本稿では，文単位のto-English言語対について議論する．
In this section, we describe the MTE metric that achieves the best performance in WMT-2017~\cite{bojar-2017} and -2018~\cite{ma-2018} Metrics Shared Task.
In this task, we use direct assessment (DA) datasets of human evaluation data.
DA datasets provides the absolute quality scores of hypotheses by measuring to what extent a hypothesis adequately expresses the meaning of the reference translation.
Each metric estimates the quality score with the translation and reference sentence pair as input, and is evaluated by Pearson correlation with human evaluation.
In this paper, we discuss the metrics task in segment-level for to-English language pairs.

\vspace{-1mm}
\subsection{Blend: the metric based on local features}
% WMT-2017において最高性能を達成したBlend\footnote{\url{http://github.com/qingsongma/blend}}~\cite{ma-2017}は，機械翻訳の自動評価用ツールキットAsiya\footnote{\url{http://asiya.lsi.upc.edu}}の基本素性に4種類の他の機械翻訳自動評価手法を組み合わせたアンサンブル手法である．
% Blendは多くの素性を用いる手法であるが，文字単位の編集距離や単語N-gramに基づく素性など，文全体を同時に考慮できない局所的な情報のみに頼っている．
Blend which achieved the best performance in WMT-2017 is an ensemble metric that incorporates 25 lexical metrics provided by the Asiya MT evaluation toolkit, as well as four other metrics.
Blend is a metric that uses many features, but relies only on local information that can not simultaneously consider the whole sentence simultaneously, such as character-based editing distances and features based on word N-grams.

\vspace{-3mm}
\subsection{RUSE: the metric based on sentence embeddings}
% WMT-2018において最高性能を達成したRUSE\footnotemark[1]~\cite{shimanaka-2018b}は，大規模コーパスによって事前学習された文の分散表現を用いる機械翻訳自動評価手法である．
% Blendなどの従来手法とは異なり，RUSEは文全体の情報を分散表現として同時に考慮できるという利点を持つ．
RUSE~\cite{shimanaka-2018b} which achieved the best performance in WMT-2018 is a metric using sentence embeddings pre-trained on large amounts of text.
Unlike previous metrics such as Blend, RUSE has the advantage of simultaneously considering the information of the whole sentence as a distributed representation.

% 文の分散表現を用いる手法には，ReVal~\cite{gupta-2015a}もある．
% ReValはWMT Metrics Shared Taskおよび文対の意味的類似度推定タスクのラベル付きデータを用いて文の分散表現を訓練するが，少量のコーパスのみを用いるため十分な性能を達成できない．
% RUSEでは，Quick Thought~\cite{logeswaran-2018}などの大規模な外部データを用いて事前学習された文の分散表現を利用し，ラベル付きデータを用いて回帰モデルのみを訓練する．
ReVal\footnote{https://github.com/rohitguptacs/ReVal}~\cite{gupta-2015a} is also a metric using sentence embeddings.
ReVal trains sentence embeddings from labeled data in WMT Metrics Shared Task and semantic similarity estimation tasks, but can not achieve sufficient performance because it uses only small data.
RUSE trains only regression models from labeled data using sentence embeddings pre-trained on large data such as Quick Thought~\cite{logeswaran-2018}.

% 図~\ref{fig:ruse}に示すように，RUSEは翻訳文と参照文を文の符号化器でそれぞれ符号化する．
% そして，InferSent~\cite{conneau-2017}にならって2つの文の分散表現を組み合わせて素性を抽出し，多層パーセプトロン(MLP)に基づく回帰モデルによって評価値を推定する．
As shown in Figure~\ref{fig:ruse}, RUSE encodes an MT hypothesis and an reference translation by a sentence encoder, respectively.
Then, following InferSent~\cite{conneau-2017}, a features are extracted by combining sentence embeddings of the two sentences, and the evaluation score is estimated by the regression model based on multi-layer perceptron (MLP).

\begin{figure*}[t]
  \centering
  \includegraphics[width=0.45\paperwidth]{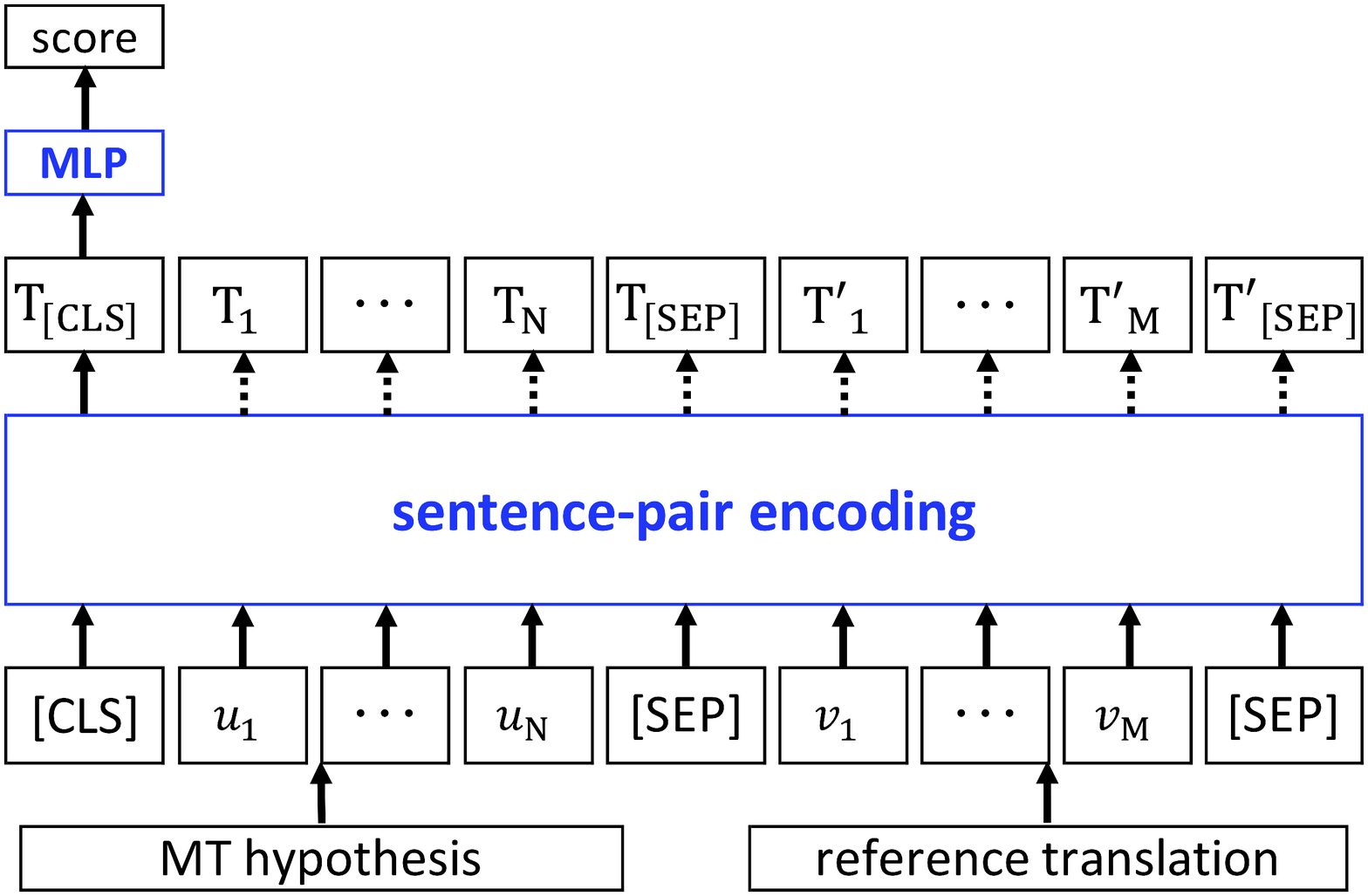}
  \caption{BERT sentence-pair encoding.}\label{fig:bert_input}
\end{figure*}

\vspace{-2mm}
\section{BERT for MTE}\label{sec:proposed_method}
% 本研究では，BERT~\cite{devlin-2019}を用いて機械翻訳の自動評価を行う．
% BERTはRUSEと同じく，事前学習された文の分散表現を利用し，MLPによって評価値を推定する．
% ただし，図~\ref{fig:bert}に示すように，BERTによる機械翻訳の自動評価では翻訳文と参照文の両方を文対の符号化器で同時に符号化する．
% そして，文対の分散表現をそのままMLPへ入力する．
% RUSEとは異なり，事前学習された符号化器もMLPとともに再訓練される．
% 以下では，BERTによる機械翻訳自動評価の特徴である事前学習の方法，文対モデリング，符号化器の再訓練について詳細に説明する．
In this study, we use BERT~\cite{devlin-2019} for MTE.
Like RUSE, BERT for MTE uses pre-trained sentence embeddings and estimates the evaluation score using the regression model based on MLP.
However, as shown in the figure~\ref{fig:bert}, in  BERT for MTE, both an MT hypothesis and an reference translation are encoded simultaneously by the sentence-pair encoder.
Then, the sentence-pair embedding is input to the regression model based on MLP.
Unlike RUSE, the pre-trained encoder is also fine-tuning with MLP.
In the following, we explain the three differences between RUSE and BERT in detail which are \textit{the pre-training method}, \textit{the sentence-pair encoding}, and \textit{the fine-tuning of the pre-trained encoder}.

\vspace{-2mm}
% \subsection{事前学習}\label{sub:bert_pre-train}
\subsection{Pre-training Method}\label{sub:bert_pre-train}
% BERTは，大規模な生コーパスを用いて，以下の2種類の教師なし事前学習を同時に行う．
BERT is designed to pre-train using two types of unsupervised task simultaneously on large amounts of raw text.

% \paragraph{双方向言語モデル}
\paragraph{Masked Language Model (MLM)}
% 生コーパスの一部のトークンを[MASK]トークンに置換した上で，双方向の言語モデルによって元のトークンを推定する．
% この教師なし事前学習によって，BERTの符号化器は文内におけるトークン間の関係を学習する．
After replacing some tokens in the raw corpus with [MASK] tokens, we estimate the original tokens by a bidirectional language model.
By this unsupervised pre-training, BERT encoder learns the relation between tokens in the sentence.

% \paragraph{隣接文推定}
\paragraph{Next Sentence Prediction (NSP)}
% 生コーパスの一部の文を無作為に他の文に置換した上で，連続する2文が隣接していた文対か否かの2値分類を行う．
% この教師なし事前学習によって，BERTの符号化器は文対の関係を学習する．
Some sentences in the raw corpus are randomly replaced with other sentences, and then binary classification is performed to determine whether two consecutive sentences are adjacent or not.
By this unsupervised pre-training, BERT encoder learns the relationship between two consecutive sentences.

% \subsection{文対モデリング}
\subsection{Sentence-pair Encoding}
% BERTでは，隣接文推定や含意関係認識などの文対を扱うタスクのために，各文を独立に符号化するのではなく，文対を同時に符号化する．
% 文対に含まれる各文は，入力系列の先頭に一度のみ追加される[CLS]トークンおよび各文末に追加される[SEP]トークンによって区別される（図~\ref{fig:bert_input}）．
% 最終的に，[CLS]トークンに対応する最終の隠れ層が，文対の分散表現を表す\footnote{極性分類などの単一文を扱うタスクのために，文対ではなく文を符号化することもできる．この場合，文頭と文末に[CLS]トークンと[SEP]トークンが一度ずつ追加され，[CLS]トークンに対応する最終の隠れ層が文の分散表現を表す．}．
In BERT, instead of encoding each sentence independently, it encodes a sentence-pairs simultaneously for task of dealing with sentence pairs such as NSP and Natural Language Inference.
The first token of every sequence is always a special classification token ([CLS]) and each sentence is separated with a special end-of-sentence token ([SEP]) (Figure~\ref{fig:bert_input}).
Finally, the final hidden state corresponding to a special [CLS] token is used as the aggregate sequence representation for classification tasks.

\vspace{-1.5mm}
% \subsection{符号化器の再訓練}
\subsection{Fine-tuning of the Pre-trained Encoder}
% BERTでは，符号化器で文または文対の分散表現を得た後，それを入力としてMLPによって分類や回帰などの応用タスクを解く．
% なお，応用タスクのラベル付きデータを用いてMLPを訓練する際，文または文対の分散表現を得るための符号化器も再訓練する．
In BERT, after obtaining a sentence embedding or a sentence-pair embedding using an encoder, it is used as an input of MLP to solve applied tasks such as classification and regression.
When training an MLP with labeled data of the applied task, we also fine-tune the pre-trained encoder.

\begin{table*}[t]
% \caption{WMT Metrics Shared Taskのto-English言語対\protect\footnotemark[5]における人手評価付き文対数}\label{tab:dataset_da}
\centering
\begin{tabular}{ccccccccc} \toprule
         & cs-en & de-en & fi-en & lv-en & ro-en & ru-en & tr-en & zh-en \\ \midrule
WMT-2015 &  500  &  500  &  500  &   -   &   -   &  500  &   -   &   -   \\
WMT-2016 &  560  &  560  &  560  &   -   &  560  &  560  &  560  &   -   \\
WMT-2017 &  560  &  560  &  560  &  560  &   -   &  560  &  560  &  560  \\ \bottomrule
\end{tabular}
\caption{Number of segment-level DA human evaluation datasets for to-English language pairs in WMT-2015~\cite{stanojevic-2015b}, WMT-2016~\cite{bojar-2016b}, and WMT-2017 Metrics Shared Task~\cite{bojar-2017}.}\label{tab:dataset_da}
\end{table*}

\begin{table*}[t]
% \caption{WMT-2017 Metrics Shared Task（文単位，to-English言語対）におけるピアソンの相関係数}\label{tab:experiment_wmt17_segment}
\centering
\begin{tabular}{lcccccccc} \toprule
                                         &      cs-en  &      de-en  &      fi-en  &      lv-en  &      ru-en  &      tr-en  &      zh-en  &      avg.   \\ \midrule
SentBLEU~\cite{bojar-2017}               &      0.435  &      0.432  &      0.571  &      0.393  &      0.484  &      0.538  &      0.512  &      0.481  \\
% chrF++~\cite{bojar-2017}                 &      0.523  &      0.534  &      0.678  &      0.520  &      0.588  &      0.614  &      0.593  &      0.579  \\
% MEANT 2.0-NOSRL~\cite{bojar-2017}      &      0.566  &      0.564  &      0.682  &      0.573  &      0.591  &      0.582  &      0.630  &      0.598  \\
% MEANT 2.0~\cite{bojar-2017}              &      0.578  &      0.565  &      0.687  &      0.586  &      0.607  &      0.596  &      0.639  &      0.608  \\
Blend~\cite{bojar-2017}                  &      0.594  &      0.571  &      0.733  &      0.577  &      0.622  &      0.671  &      0.661  &      0.633  \\
RUSE~\cite{shimanaka-2018b}    &      0.614  &      0.637  &      0.756  &      0.705  &      0.680  &      0.704  &      0.677  &      0.682  \\
% RUSE with BERT$_\text{BASE[SEP]}$        &      0.645  &      0.612  &      0.759  &      0.724  &      0.615  &      0.689  &      0.618  &      0.666  \\
% RUSE with BERT$_\text{LARGE[SEP]}$       &      0.607  &      0.600  &      0.755  &      0.667  &      0.641  &      0.658  &      0.602  &      0.647  \\
% RUSE with BERT$_\text{BASE}$ (sent-pair) &      0.645  &      0.607  &      0.780  &      0.727  &      0.644  &      0.704  &      0.705  &      0.687  \\
% RUSE with BERT$_\text{LARGE}$ (sent-pair)&      0.645  &      0.607  &      0.780  &      0.727  &      0.644  &      0.704  &      0.687  &      0.669  \\
BERT$_\text{BASE}$                       & {\bf 0.720} & {\bf 0.761} & {\bf 0.857} & {\bf 0.828} & {\bf 0.788} & {\bf 0.798} & {\bf 0.763} & {\bf 0.788} \\ \bottomrule
% BERT$_\text{LARGE}$                      & {\bf 0.727} & {\bf 0.767} & {\bf 0.848} & {\bf 0.828} & {\bf 0.794} & {\bf 0.807} & {\bf 0.761} & {\bf 0.790} \\
\end{tabular}
\caption{Segment-level Pearson correlation of metric scores and DA human evaluation scores for to-English language pairs in WMT-2017 Metrics Shared Task.}\label{tab:experiment_wmt17_segment}
\end{table*}

\vspace{-1.5mm}
% \section{評価実験}\label{sec:experiment}
\section{Experiments}\label{sec:experiment}
% WMT-2017 Metrics Shared Task~\cite{bojar-2017}のデータセットを用いて，文単位のto-English言語対におけるBERTの有効性を検証する．
We performed experiments using the WMT-2017 Metrics Shared Task dataset to verify the performance of BERT for MTE.

\vspace{-1.5mm}
\subsection{Settings}\label{sub:setting}
% 表~\ref{tab:dataset_da}に，データセットの文対数を示す．
% WMT-2015およびWMT-2016の合計5,360文対は無作為に分割し，9割を訓練用，1割を開発用に利用する．
% WMT-2017の文対は評価用に利用する．
Table~\ref{tab:dataset_da} shows the number of instances in WMT Metrics Shared Task dataset (segment-level) for to-English language pairs\footnote{en: English, cs: Czech, de: German, fi: Finnish, ro: Romanian, ru: Russian, tr: Turkish, lv: Latvian, zh: Chinese} used in this study.
A total of 5,360 instances in WMT-2015 and WMT-2016 Metrics Shared Task datasets will be divided randomly, and 90\% is used for training and 10\% for development.
A total of 3,920 instances (560 instances for each language pair) in WMT-2017 Metrics Shared Task dataset is used for evaluation.

% 比較手法には，WMT Metrics Shared TaskのベースラインであるSentBLEU~\cite{bojar-2017}，WMT-2017にて最高性能を達成したBlend~\cite{ma-2017}，WMT-2018にて最高性能を達成したRUSE~\cite{shimanaka-2018b}を用いる．
% 各手法は，ピアソンの相関係数を用いて人手評価との相関を評価される．
As a comparison method, we use SentBLEU\footnote{\url{https://github.com/moses-smt/mosesdecoder/blob/master/scripts/generic/mteval-v13a.pl}} which is the baseline of WMT Metrics Shared Task, Blend~\cite{ma-2017} which achieved the best performance in WMT-2017 Metrics Shared Task, and RUSE~\cite{shimanaka-2018b} which achieved the best performance in WMT-2018 Metrics Shared Task.
We evaluated each metric using the Pearson correlation coefficient between the metric scores and the DA human scores.

% \footnotetext[5]{en: English, cs: Czech, de: German, fi: Finnish, ro: Romanian, ru: Russian, tr: Turkish, lv: Latvian, zh: Chinese}
% \addtocounter{footnote}{+1}

% BERTには，著者らによって公開されている訓練済みモデルのうち，BERT$_\text{BASE}$~(uncased)\footnote{\url{https://github.com/google-research/bert}}を用いる．
% BERTのパラメータは，以下の組み合わせの中からグリッドサーチによって選択する．
Among the trained models published by the authors, BERT$_\text{BASE}$~(uncased)\footnote{\url{https://github.com/google-research/bert}} is used for MTE with BERT.
The Hyper-parameters for fine-tuning BERT are determined through grid search in the following parameters using the development data.

\begin{itemize}
\item \begin{tabbing}
Aaaaaaaaaaaaaaaa \=   a   \=    aaaaaaaaa \kill
$\text{Batch~size} \in \{16, 32\}$ \end{tabbing}
\item \begin{tabbing}
Aaaaaaaaaaaaaaaa \=   a   \=    aaaaaaaaa \kill
$\text{Learning~rate} (\text{Adam}) \in \{\text{5e-5}, \text{3e-5}, \text{2e-5}\}$ \end{tabbing}
\item \begin{tabbing}
Aaaaaaaaaaaaaaaa \=   a   \=    aaaaaaaaa \kill
$\text{Number~of~epochs} \in \{3, 4\}$ \end{tabbing}
\item \begin{tabbing}
Aaaaaaaaaaaaaaaa \=   a   \=    aaaaaaaaa \kill
$\text{Dropout~rate (MLP)} \in \{0.1\}$ \end{tabbing}
\item \begin{tabbing}
Aaaaaaaaaaaaaaaa \=   a   \=    aaaaaaaaa \kill
$\text{Number~of~hidden~layers (MLP)} \in \{0\}$ \end{tabbing}
\item \begin{tabbing}
Aaaaaaaaaaaaaaaa \=   a   \=    aaaaaaaaa \kill
$\text{Number~of~hidden~units (MLP)} \in \{768\}$ \end{tabbing}
\end{itemize}

% \begin{itemize}
%  \item Batch~size \in \{16, 32\}
%  \vspace{-1.5mm}
%  \item Learning~rate（\text{Adam}） \in \{\text{5e-5}, \text{3e-5}, \text{2e-5}\}
%  \vspace{-1.5mm}
%  \item Number~of~epochs \in \{3, 4\}
%  \vspace{-1.5mm}
%  \item Dropout~rate \in \{0.1\}
%  \vspace{-1.5mm}
%  \item \text{Number~of~layers} \in \{0\}
%  \vspace{-1.5mm}
%  \item \text{Number~of~units} \in \{768\}
%  \vspace{-1.5mm}
% \end{itemize}

\begin{table*}[t]
\centering
\begin{tabular}{lcccccccc} \toprule
                                         &      cs-en  &      de-en  &      fi-en  &      lv-en  &      ru-en  &      tr-en  &      zh-en  &      avg.   \\ \midrule
RUSE with GloVe-BoW                      &      0.475  &     0.479   &      0.645  &      0.532  &      0.537  &      0.547  &      0.480  &      0.527  \\
% RUSE with IS                             &      0.556  &      0.568  &      0.706  &      0.650  &      0.626  &      0.649  &      0.634  &      0.627  \\
RUSE with Quick Thought                  &      0.599  &      0.588  &      0.736  &      0.690  &      0.655  &      0.710  &      0.645  &      0.660  \\
% RUSE with USE                            &      0.592  &      0.596  &      0.681  &      0.621  &      0.598  &      0.645  &      0.620  &      0.622  \\
RUSE with BERT                           &      0.622  &      0.626  &      0.765  &      0.708  &      0.609  &      0.706  &      0.647  &      0.669  \\
% RUSE with BERT$_\text{LARGE}$ (sent)     &      0.669  &      0.637  &      0.756  &      0.705  &      0.679  &      0.716  &      0.666  &      0.690  \\
% RUSE with IS+QT+USE~\cite{shimanaka-2018b}    &      0.614  &      0.637  &      0.756  &      0.705  &      0.680  &      0.704  &      0.677  &      0.682  \\
% RUSE with BERT$_\text{BASE[SEP]}$        &      0.645  &      0.612  &      0.759  &      0.724  &      0.615  &      0.689  &      0.618  &      0.666  \\
% RUSE with BERT$_\text{LARGE[SEP]}$       &      0.607  &      0.600  &      0.755  &      0.667  &      0.641  &      0.658  &      0.602  &      0.647  \\
BERT (w/o fine-tuning)                   &      0.645  &      0.607  &      0.780  &      0.727  &      0.644  &      0.704  &      0.705  &      0.687  \\
% RUSE with BERT$_\text{LARGE}$ (sent-pair)&      0.645  &      0.607  &      0.780  &      0.727  &      0.644  &      0.704  &      0.687  &      0.669  \\
BERT                                     & {\bf 0.720} & {\bf 0.761} & {\bf 0.857} & {\bf 0.828} & {\bf 0.788} & {\bf 0.798} & {\bf 0.763} & {\bf 0.788} \\ \bottomrule
% BERT$_\text{LARGE}$                      & {\bf 0.727} & {\bf 0.767} & {\bf 0.848} & {\bf 0.828} & {\bf 0.794} & {\bf 0.807} & {\bf 0.761} & {\bf 0.790} \\
\end{tabular}
\caption{Comparison of RUSE and BERT in WMT-2017 Metrics Shared Task (segment-level, to-English language pairs).}\label{tab:analysis}
\end{table*}

\vspace{-2mm}
\subsection{Results}\label{sub:result}

% 表~\ref{tab:experiment_wmt17_segment}に実験結果を示す．
% BERTは，全ての言語対で他の手法を大幅に上回る性能を達成した．
% \ref{sec:discussion}節では，RUSEとBERTを比較しつつ，詳細な分析を行う．
Table~\ref{tab:experiment_wmt17_segment} presents the experimental results of the WMT-2017 Metrics Shared Task dataset.
BERT for MTE achieved the best per-formance in all to-English language pairs.
In Section~\ref{sec:discussion}, we compare RUSE and BERT and do a detailed analysis.

\vspace{-2mm}
\section{Analysis: Comparison of RUSE and BERT}\label{sec:discussion}
% RUSEとBERTの主な相違点である{\bf 事前学習の方法}，{\bf 文対モデリング}，{\bf 符号化器の再訓練}の3点に関して分析するために，以下の設定で実験を行う．
In order to analyze the three main points of difference between RUSE and BERT, \textit{the pre-training method}, \textit{the sentence-pair encoding}, and \textit{the fine-tuning of the pre-trained encoder}, we conduct an experiment with the following settings.

% \begin{itemize}
%   \vspace{-1.5mm}
%   \item RUSE with GloVe-BoW: 図~\ref{fig:ruse}の文の分散表現として，単語分散表現GloVe~\cite{pennington-2014} (glove.840B.300d\footnote{\url{https://nlp.stanford.edu/projects/glove}})の平均ベクトルを用いる．
%   \vspace{-1.5mm}
%   \item RUSE with Quick Thought: 図~\ref{fig:ruse}の文の符号化器として，隣接文推定によって事前学習されたQuick-Thought~\cite{logeswaran-2018}を用いる．
%   \vspace{-1.5mm}
%   \item RUSE with BERT$_\text{BASE}$（文）: 図~\ref{fig:ruse}の文の符号化器として，双方向言語モデルと隣接文推定によって事前学習された単一文入力のBERTを用いる．ただし，文の符号化器は再訓練しない．
%   \vspace{-1.5mm}
%   \item RUSE with BERT$_\text{BASE}$（文対）: 図~\ref{fig:ruse}のMLPの入力として，文対を入力とするBERTの出力を用いる．ただし，文対の符号化器は再訓練しない．
%   \vspace{-1.5mm}
% \end{itemize}

\paragraph{RUSE with GloVe-BoW: }
% Figure~\ref{fig:ruse}の文の分散表現として，単語分散表現GloVe~\shortcite{pennington-2014} (glove.840B.300d\footnote{\url{https://nlp.stanford.edu/projects/glove}})の平均ベクトルを用いる．
The mean vector of word embeddings of GloVe~\cite{pennington-2014}(glove.840B.300d\footnote{\url{https://nlp.stanford.edu/projects/glove}}) (300 dimension) in each sentence is used as the sentence embeddings in Figure~\ref{fig:ruse}.

\paragraph{RUSE with Quick Thought: }
% BookCorpusデータセット~\shortcite{Zhu-2015}の4,500万文およびUMBC WebBase~\shortcite{han-2013}の約1億3,000万文の両方を用いて事前学習されたQuick Thoughtによって4,800次元の文の分散表現を獲得し，\ref{subsub:ruse_regressor}節の方法で素性を抽出する．
Quick Thought~\cite{logeswaran-2018} pre-trained on both 45 million sentences in the BookCorpus~\cite{Zhu-2015} and about 130 million sentences in UMBC WebBase coupus~\cite{han-2013} is used as the sentence encoder in Figure~\ref{fig:ruse}.

\paragraph{RUSE with BERT: }
% 単一文を入力とするBERTの[CLS]トークンに対応する隠れ層のうち，最終4層を連結したものを3,072次元の文の分散表現として\ref{subsub:ruse_regressor}節の方法で素性を抽出する．
% ただし，BERTの符号化器の部分は再訓練しない．
A concatenation of the last four hidden layers (3,072 dimention) corresponding to the [CLS] token of BERT that takes a single sentence as input is used as the sentence embeddings in Figure~\ref{fig:ruse}.

\paragraph{BERT (w/o fine-tuning): }
% 文対を入力とするBERTの[CLS]トークンに対応する隠れ層のうち最終4層を連結したもの（3,072次元）を，図~\ref{fig:bert}のMLPの入力として用いる．
% ただし，BERTの符号化器の部分は再訓練しない．
A concatenation of the last four hidden layers (3,072 dimension) corresponding to the [CLS] token of BERT that takes a sentence-pair as the input sequence is used as the input of the MLP Regressor in Figure~\ref{fig:bert}.
In this case, the part of the BERT encoder is not fine-tuned.

\paragraph{BERT: }
% 文対を入力とするBERTの[CLS]トークンに対応する最終隠れ層（768次元）を図~\ref{fig:bert}のMLPの入力として用い，MLPとともにBERTの符号化器の部分も再訓練する．
The last hidden layer (768 dimension) corresponding to the [CLS] token of BERT that takes a sentence-pair as the input sequence is used as the input of the MLP Regressor in Figure~\ref{fig:bert}.
In this case, the part of the BERT encoder is fine-tuned.

% RUSEの素性としてBERTを用いる場合，[CLS]トークンに対応する隠れ層のうち最終4層を連結し，文または文対の分散表現として用いる．
% RUSEのパラメータは先行研究~\cite{shimanaka-2018b}にならって，以下の組み合わせの中からグリッドサーチによって選択する．
% RUSEとBERT (w/o fine-tuning)の各パラメータは，以下の組み合わせの中からグリッドサーチにより選択する．
The Hyper-parameters for RUSE and BERT (w/o fine-tuning) are determined through grid search in the following parameters using the development data.

% \begin{itemize}
%  \item $バッチサイズ \in \{64, 128, 256, 512, 1024\}$
%  \vspace{-1.5mm}
%  \item $学習率（\text{Adam}） \in \{\text{1e-3}\}$
%  \vspace{-1.5mm}
%  \item $エポック数 \in \{1, 2, ..., 30\}$
%  \vspace{-1.5mm}
%  \item $\text{ドロップアウト率} \in \{0.1, 0.3, 0.5\}$
%  \vspace{-1.5mm}
%  \item $\text{MLPの隠れ層の数} \in \{1, 2, 3\}$
%  \vspace{-1.5mm}
%  \item $\text{MLPの隠れ層の次元} \in \{512, 1024, 2048, 4096\}$
% \end{itemize}

\begin{itemize}
\item \begin{tabbing}
Aaaaaaaaaaaaaaaa \=   a   \=    aaaaaaaaa \kill
$\text{Batch~size} \in \{64, 128, 256, 512, 1024\}$ \end{tabbing}
\item \begin{tabbing}
Aaaaaaaaaaaaaaaa \=   a   \=    aaaaaaaaa \kill
$\text{Learning~rate} (\text{Adam}) \in \{\text{1e-3}\}$ \end{tabbing}
\item \begin{tabbing}
Aaaaaaaaaaaaaaaa \=   a   \=    aaaaaaaaa \kill
$\text{Number~of~epochs} \in \{1, 2, ..., 30\}$ \end{tabbing}
\item \begin{tabbing}
Aaaaaaaaaaaaaaaa \=   a   \=    aaaaaaaaa \kill
$\text{Dropout~rate (MLP)} \in \{0.1, 0.3, 0.5\}$ \end{tabbing}
\item \begin{tabbing}
Aaaaaaaaaaaaaaaa \=   a   \=    aaaaaaaaa \kill
$\text{Number~of~hidden~layers (MLP)} \in \{1, 2, 3\}$ \end{tabbing}
\item \begin{tabbing}
Aaaaaaaaaaaaaaaa \=   a   \=    aaaaaaaaa \kill
$\text{Number~of~hidden~units (MLP)} \in \{512, 1024, 2048, 4096\}$ \end{tabbing}
\end{itemize}

% これらの実験結果を表~\ref{tab:analysis}に示す．
Table~\ref{tab:analysis} presents these experimental results of the WMT-2017 Metrics Shared Task dataset.

\paragraph{Pre-training Method}
% 表~\ref{tab:analysis}の上から3行を比較すると，文の符号化器における事前学習の方法による性能への影響がわかる．
% まず，単語の分散表現に基づくGloVe-BoWよりも，文の分散表現に基づくQuick Thoughtの方が，一貫して高い性能を持つ．
% 次に，隣接文推定のみによって事前学習されたQuick Thoughtよりも，双方向言語モデルと隣接文推定の両方によって事前学習されたBERT$_\text{BASE}$（文）の方が，多くの言語対において優れた性能を発揮する．
% つまり，BERTの大きな特徴のひとつである双方向言語モデルによる事前学習は，機械翻訳の自動評価のためにも有用である．
The top three rows of Table~\ref{tab:analysis} show the performance impact of the method of pre-learning in the sentence encoder.
First, Quick Thought based on sentence embeddings has better performance consistently than GloVe-BoW based on word embeddings.
Second, BERT pret-rained by both MLM and NSP perform better on many language pairs than Quick Thought pre-trained only by NSP.
In other words, the pre-training method using Masked Language Model (MLM), which is one of the major features of BERT, is also useful for MTE.

\vspace{-1mm}
\paragraph{Sentence-pair Encoding}
% RUSE with BERT$_\text{BASE}$（文）とRUSE with BERT$_\text{BASE}$（文対）を比較すると，文対モデリングによる性能への影響がわかる．
% 多くの言語対において，翻訳文と参照文を独立に符号化する前者よりも，同時に符号化する後者の方が高い性能を持つ．
% RUSEではInferSentにならって2つの文の分散表現を組み合わせる素性抽出を行っているが，これが機械翻訳の自動評価に適した素性抽出の方法であるとは限らない．
% 一方で，BERTの文対モデリングは，素性抽出を陽に行うことなく文対の関係を考慮した分散表現を得ている．
% BERTでは，隣接文推定による事前学習の際に，上手く文対の関係を学習できている可能性がある．
Comparing RUSE with BERT and BERT (w/o fine-tuning) shows the impact of the sentence-pair encoding on the performance of MTE.
In the case of many language pairs, the latter, which simultaneously encodes an MT hypothesis and a reference translation, has higher performance than the former, which encodes them independently.
Although RUSE performs feature extraction that combines sentence embeddings of two sentences in the same way as InferSent, this is not necessarily the method of feature extraction suitable for MTE.
On the other hand, the sentence-pair encoding of BERT obtains sentence embeddings considering the relation of sentence-pair without explicitly extracting the feature.
In BERT, there is a possibility that the relation of sentence-pair can be trained well at the time of pre-training by NSP.

\vspace{-1mm}
\paragraph{Fine-tuning of the Pre-trained Encoder}
% 表~\ref{tab:analysis}の下から2行を比較すると，符号化器の再訓練による性能への影響がわかる．
% 全ての言語対において，MLPのみを訓練するRUSE with BERT$_\text{BASE}$（文対）よりも，MLPとともに符号化器を再訓練するBERT$_\text{BASE}$の方が大幅に優れた性能を発揮する．
% つまり，BERTの大きな特徴のひとつである符号化器の再訓練は，機械翻訳の自動評価のためにも有用である．
The bottom two rows of Table~\ref{tab:analysis} show the performance impact of the fine-tuning of the pre-trained encoder.
In the case of all language pairs, BERT, which fine-tune the pre-trained encoder with MLP, performs much better than RUSE, which only trains MLP.
In other words, the fine-tuning of the pre-trained encoder, which is one of the major features of BERT, is also useful for machine translation evaluation.

% \section{おわりに}\label{sec:outro}
\section{Conclusion}\label{sec:outro}
% 本研究では，BERTを用いた機械翻訳の自動評価を行った．
% 実験の結果，BERTは文単位の全てのto-English言語対で他の手法を大幅に上回り，最高性能を更新した．
% また，先行研究のRUSEとの比較に基づく分析の結果，事前学習の方法，文対モデリング，符号化器の再訓練の3つの点が，それぞれBERTの性能改善に貢献していることを示した．
In this study, we proposed the metric for automatic machine translation evaluation with BERT.
Our segment-level MTE metric with BERT achieved the best performance in segment-level metrics tasks on the WMT17 dataset for all to-English language pairs.
In addition, as a result of analysis based on comparison with RUSE which is our previous work, it is shown that three points of \textit{the pre-training method}, \textit{the sentence-pair encoding}, and \textit{the fine-tuning of the pre-trained encoder} contributed to the performance improvement of BERT respectively.

\small
\paragraph{Acknowledgement}
% 本研究の一部はJSPS科研費（研究活動スタート支援，課題番号: 18H06465）の助成を受けたものです．
Part of this research was funded by JSPS Grant-in-Aid for Scientific Research (Grant-in-Aid for Research Activity start-up, task number: 18H06465).
\vspace{-3mm}

% References
\bibliographystyle{acl_natbib_nourl}
\bibliography{arXiv2019}
\end{document}